\documentclass[journal]{IEEEtran}
\usepackage{graphicx}
\usepackage{amsmath}
\usepackage{times}
\usepackage{epsfig}
\usepackage{amssymb}
\usepackage{epstopdf}
\usepackage{float}
\usepackage{multirow}
\usepackage{caption}

\def\normalsize{\small}

\ifCLASSINFOpdf
\else
\fi
\begin{document}
\title{Temporal Pyramid Pooling Based Convolutional Neural Network for Action Recognition}

\author{Peng Wang,
\thanks{P. Wang's contribution was made when visiting the University of Adelaide.
The first two authors made equal contributions to this work.}
        Yuanzhouhan Cao,
        Chunhua Shen,
        Lingqiao Liu,
        and Heng Tao Shen%
\thanks{P. Wang and H. T. Shen are with School of Information Technology and Electrical Engineering,
  The University of Queensland, Australia (email: p.wang6@uq.edu.au;
shenht@itee.uq.edu.au).}%
\thanks{Y. Cao, C. Shen, and  L. Liu are with School of Computer Science, The University of Adelaide, Australia
(email: \{yuanzhouhan.cao, chunhua.shen, lingqiao.liu\}@adelaide.edu.au).}

\thanks{C. Shen is also with Australian Centre for Robotic Vision, Australia.}%
}

\markboth{Temporal Pyramid Pooling CNN for Action Recognition}%
{Wang et al.}
\maketitle

\begin{abstract}
Encouraged by the success of Convolutional Neural Networks (CNNs) in image
classification, recently much effort is spent on applying CNNs
to video based action recognition problems. One challenge is that video
contains a varying number of frames which is incompatible to the standard input
format of CNNs. Existing methods handle this issue either by directly sampling
a fixed number of frames or bypassing this issue by introducing a 3D
convolutional layer which conducts convolution in spatial-temporal domain.

In this paper we propose a novel network structure which allows an
arbitrary number of frames as the network input. The key of our solution is to
introduce a module consisting of an encoding layer and a temporal pyramid
pooling layer. The encoding layer maps the activation from previous layers to a
feature vector suitable for pooling while the temporal pyramid pooling layer
converts multiple frame-level activations into a fixed-length video-level
representation. In addition, we adopt a feature concatenation layer which
combines appearance information and motion information. Compared with the frame
sampling strategy, our method avoids the risk of missing any important frames.
Compared with the 3D convolutional method which requires a huge video dataset
for network training, our model can be learned on a small target dataset
because we can leverage the off-the-shelf image-level CNN for model parameter
initialization. Experiments on two challenging datasets, Hollywood2 and HMDB51,
demonstrate that our method achieves superior performance over state-of-the-art
methods while requiring much fewer training data.
\end{abstract}


\IEEEpeerreviewmaketitle

\section{Introduction}

\begin{figure}[h]
\begin{center}
\includegraphics[scale=.5]{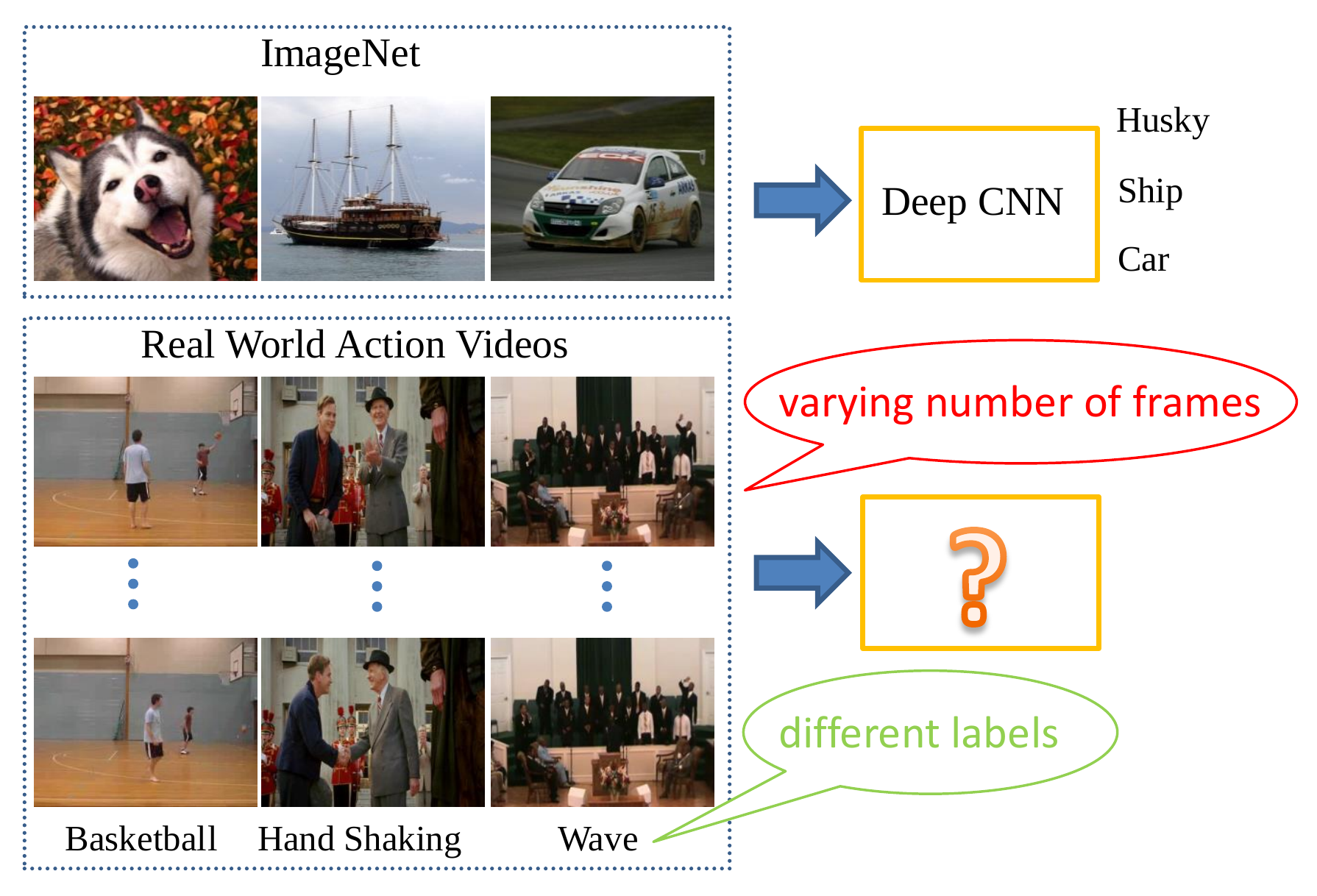}
\end{center}
   \caption{Image-level CNN requires a single image as input while videos have varying number of frames.}
\label{fig:network}
\end{figure}
\IEEEPARstart{H}ow to design a feature representation to fully exploit
the spatial-temporal information in videos constitutes a cornerstone in video
based human action recognition. Current state-of-the-art methods usually
generate video-level representations by adopting hand-crafted features such as
spatial-temporal interest points \cite{4587756} or trajectories
\cite{wang:2011, Wang2013} and unsupervised feature encoding methods such as
Fisher vector encoding \cite{Perronnin:2010}.

Recently, deep Convolutional Neural Networks has been established as the state-of-the-art method in image classification \cite{NIPS2012_4824} and it has been demonstrated that a CNN pretrained on a large image dataset,
such as ImageNet \cite{imagenet_cvpr09}, can be used to initialize networks built for other visual recognition tasks.
Inspired by the success of CNNs in image recognition, some studies attempt to apply CNNs to video based action recognition.
However, most existing deep models are designed to work with single image input. It is non-trivial to
extend these models to videos since video clips often contain a varying number of frames.
To handle this problem, the work in \cite{Andrew14} samples a fixed number of frames and reshapes
them into a compatible input format of an image-based CNN. However, sampling may  risk  missing
important frames for action recognition, especially in videos with uncontrolled scene variation.
Another strategy is to bypass this issue by directly using videos as input and replacing the 2D convolution with 3D convolution which is operated on the spatial-temporal domain. However, the above strategy sacrifices the possibility of leveraging the powerful off-the-shelf image-based CNN to initialize model parameters or extract mid-level features. Thus, it has to rely on a huge number of training videos to avoid the risk of over-fitting. For example, the authors in \cite{KarpathyCVPR14} collect a dataset of 1 million YouTube videos for network training which takes weeks to train with modern GPUs.

In this work, we propose a novel network structure which allows an arbitrary number of video frames as input.
This is achieved by designing a module which consists of an encoding layer and a temporal pyramid pooling layer.
The encoding layer maps the activations from previous layer to a feature vector suitable for pooling,
which is akin to the encoding operation in the traditional bag-of-features model.
The temporal pyramid pooling layer converts multiple frame-level activations into a fixed-length video-level representation.
At the same time, the temporal pyramid pooling layer explicitly considers the weak temporal structure within videos.
In addition, we also introduce a feature concatenation layer into our network to combine motion and appearance information.

\section{Related work} \label{related work} Our method is related to a large
body of works on creating video representations for action recognition. Most
existing methods rely on hand-crafted  {\em shallow} features, for example, the
  sparse spatial-temporal interest points \cite{4587756},  sparse trajectories
  \cite{5459154, 5206721} and local dense trajectories
  \cite{wang:2011,Wang2013}. Usually unsupervised encoding such as the
  bag-of-features model \cite{4587756} or Fisher vector encoding
  \cite{Perronnin:2010} are applied to aggregate information from local
  descriptors into a video-level representation.

In terms of local feature descriptors, the dense trajectory has received much
attention since it has significantly boosted action recognition accuracy
\cite{wang:2011,Wang2013}.  Different from previous methods, it tracks densely
sampled points using dense optical flow. To compensate for camera motion, the
motion boundary histograms (MBH) \cite{Dalal:2006} are employed as motion
descriptors which are more discriminative than optical flow for action
recognition. In order to further improve the performance of dense trajectory,
Wang \textit{et al.} \cite{Wang2013} conduct video stabilization to remove
camera motion and use Fisher vector \cite{Perronnin:2010} to encode trajectory
descriptors. There are also works researching the fusion strategies of different descriptors of dense trajectories, e.g., HOG, HOF and MBH. In \cite{peng14}, the authors conclude that the encoding-level fusion performs better comparing to descriptor-level and classification-level fusions. And in \cite{6909477}, the authors map different descriptors into a common space to fully utilize the correlation between them. Also, it is claimed \cite{peng14} that combining multiple bag-of-features models of different descriptors can further boost the
performance.

Another type of works achieve action recognition via mining discriminative mid-level representations such as subvolumes \cite{peng:stack}, attributes \cite{NIPS2014_5565}, action parts \cite{DBLP:conf/mm/LiangLC13}, salient regions \cite{6751447} or actons \cite{6751554}. Some methods train a classifier for each discriminative part and fuse the classification scores to get a video-level output. Some other methods treat the mid-level representations as local features and encode them again using strategies, like the Fisher Vector, to derive the global representation.

Apart from the aforementioned shallow representation based methods, deep models are
also investigated for action recognition. In \cite{6165309}, Ji \textit{et al.}
propose a 3D CNN model which performs 3D convolution over spatial-temporal
domain to capture motion information encoded in consecutive frames. To avoid
the over-fitting problem in spatial-temporal model, Karpathy \textit{et al.}
\cite{KarpathyCVPR14} collect a dataset of one million video clips for network training.
They also compare several fusion methods to evaluate their effectiveness in
capturing spatial-temporal information. To speed up the training process, they
separate the architecture into two streams: one stream captures the
high-frequency detail of an object of interest from high-resolution center crop
and the other stream captures the context information from low-resolution frames. In \cite{arXiv:1412.0767} the authors collect another large-scale video dataset and propose a generic spatial-temporal features for video analysis based on 3D convolution. To better utilize the knowledge of images such as ImageNet to boost video classification performance, the authors in \cite{arXiv:1503.07274} propose several strategies to initialize the weights in 3D convolutional layers using the weights learned from 2D images.

Recently, a two-stream deep model is proposed in \cite{Andrew14} for
action recognition. While the spatial stream adopts an image-based network
structure to extract appearance features from still images, the temporal stream
is based on dense optical flow extracted from multiple frames to capture motion
information. They conduct score-level lazy fusion of these two streams to obtain
the final recognition score.

\section{Our proposed network architecture}
\label{network}
\subsection{Network overview}

\begin{figure*}[ht]
\begin{center}
\includegraphics[scale=.7]{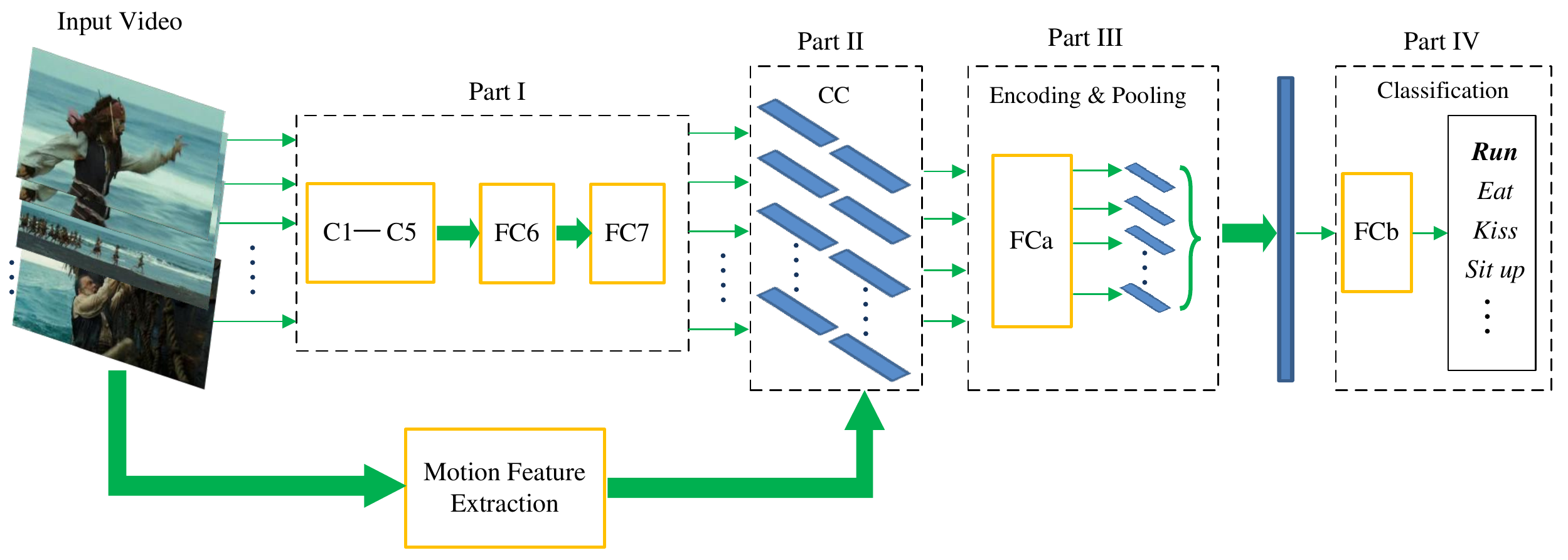}
\end{center}
   \caption{Overall structure of the proposed network. We extract appearance and motion representation from a video and concatenate them. After encoding and temporal pooling, we get the final representation of the video for classification.}
\label{fig:network}
\end{figure*}

The overall structure of our proposed network is shown in Figure \ref{fig:network}. It can be decomposed into four parts: the first part consists of five convolutional layers C1-C5 and two fully-connected layers FC6 and FC7. In the second part, the activation of FC7 is fed into the feature concatenation layer CC, which concatenates the output of FC7 and the frame-level motion feature. The concatenated feature then passes through a fully-connected layer FCa followed by a temporal pyramid pooling layer which converts frame-level features into the video-level feature. FCa together with the temporal pyramid pooling layer constitute the third part of our network, which is also the core part of our network. Finally, the classification layer, which is the fourth part of the network, is applied to the video-level feature to obtain the classification result. In the following sections, we discuss these four parts in detail.

\subsection{Network architecture}
\subsubsection{Part I:  C1 - FC7}

The first part of our network is used to generate the frame-level appearance feature. We choose the structure of this part to be the same as an off-the-shelf CNN model. Thus, we can take advanatage of the model parameters pretrained on a large dataset, e.g., ImageNet \cite{imagenet_cvpr09} to initialize our network. More specifically, this part comprises 5 convolutional layers and 2 fully connected layers. The first convolutional layer has 64 kernels of size $11\times11\times3$ and a stride of 4 pixels which ensures fast processing speed. The second convolutional layer has 256 kernels of size $5\times5\times3$. The third, forth and fifth convolutional layers have 256 kernels of size $3\times3\times3$. Two fully connected layers both have 4096 neurons. Each frame in an input video is first resized to $224\times224\times3$ and then passes through the first part of our network, interleaved with ReLU non-linearties, max-pooling and normalization. The output of the seventh layer, a 4096 dimensional vector is then used as the static appearance feature of a video frame. At the training stage, we initialize the parameters of these seven layers with a pre-trained network in \cite{Chatfield14}.

\subsubsection{Part II:  Feature concatenation layer and frame-level motion features}
\label{motion_featue}
\label{sect:concatenation}
\begin{figure}[t]
\begin{center}
   \includegraphics[width=\linewidth]{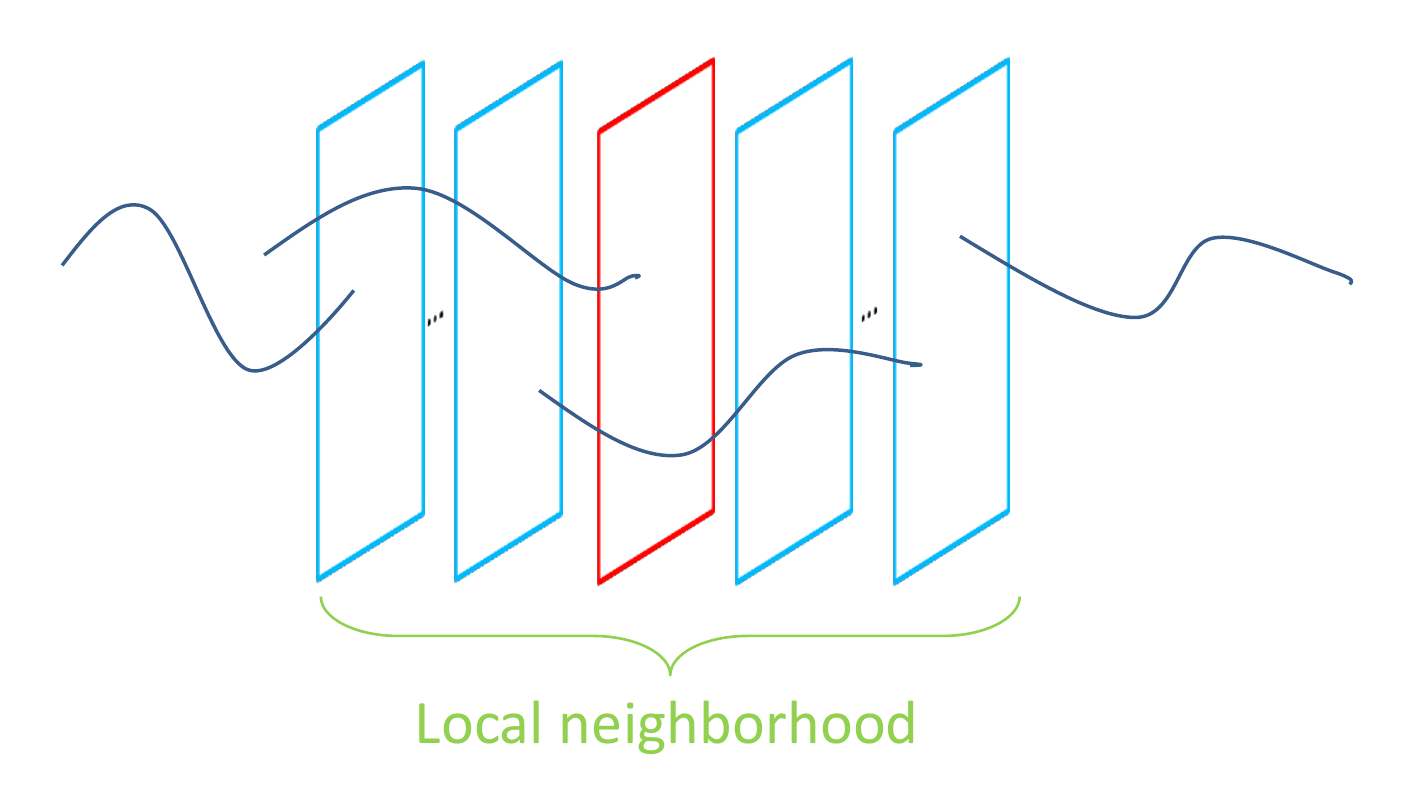}
\end{center}
   \caption{Illustration of frame-level motion feature generation. We extract the motion feature for the frame with red border. All the trajectories passing a local neighbourhood will be considered.}
\label{fig:frame-motion}
\end{figure}

We introduce a feature concatenation layer to combine appearance and motion features since they have been shown to compensate each other in representing actions in videos. Our motion feature is built upon the dense trajectory descriptor because it achieves state-of-the-art performance. We only use HOF and MBH descriptors of improved dense trajectory \cite{Wang2013} to describe motion information since we find that the 30-dimensional Trajectory descriptor in \cite{Wang2013} does not contribute too much to the classification performance. Also, instead of utilizing trajectory features to describe motion information of the whole video, we extract motion features from a short temporal range, that is, within several consecutive frames. Figure \ref{fig:frame-motion} illustrates this idea. For each frame, we extract the trajectories passing through a local neighbourhood and encode them using the Fisher Vector. The motion feature of this frame is obtained by pooling all the Fisher vector encodings within this neighbourhood. Then this motion feature is concatenated with the appearance feature from FC7 to produce the frame-level representation. We refer to this fusion method as ``early fusion".

In practice, however, the dimensionality of the Fisher vector encoding is too high to be handled by our network implemented on GPU. Thus, we employ a supervised feature merging algorithm variant in \cite{Liu:2013} (Eq. (7) in \cite{Liu:2013}) to perform dimensionality reduction. Compared with other methods, this method is very efficient in learning the dimensionality reduction function and performing dimensionality reduction especially in the scenario of reducing high-dimensional features to medium dimensions. More specifically, one only needs to calculate the mean of features in each class, which gives a data matrix $\mathbf{\bar S} \in \mathbb{R}^{c\times D}$, where $D$ indicates the feature dimensionality and $C$ indicates the total number of classes. Each column of $\mathbf{\bar S}$, denoted as $\mathbf{s}_i,~i= 1,\cdots,D$, is treated as a $c$-dimensional `signature' for the $i$-th feature. Then we perform $k$-means clustering on all $D$ `signatures' to group them into $k$ clusters. Thus the $D$ feature dimensions are partitioned into $k$ groups and this grouping pattern is used to perform dimensionality reduction. The details are illustrated in Algorithm 1.

\begin{table}[t]
\centering
\resizebox{.5\textwidth}{!} {
\renewcommand{\arraystretch}{1.25}
\begin{tabular}{l}
\hline\noalign{\smallskip}
\textbf{Algorithm 1} Feature Merging Based Dimensionality Reduction \\
\hline\noalign{\smallskip}
   1: Given a feature matrix $V\in\mathbb{R}^{m\times{D}}$, with $m$ the number of \\ \quad training samples, $D$ the dimension of FV representations.\\
   2: For $V$, merging the features belonging to same class via\\ \quad average pooling, we get $\mathbf{\bar S} \in \mathbb{R}^{c\times{D}}$, with $c$ the number of \\ \quad classes.\\
   3: Conduct k-means on columns of $\mathbf{\bar S}$, with number of clusters \\
\quad  $k$ being the target low dimension to which we want to \\ \quad reduce $D$. \\
   4: We define $clique_j = \{i|cluster(i)=j\}$,\\ \quad $norm_j=\sqrt{|\{i|cluster(i)=j\}|}$, \\ \quad with $i$ the feature indices, $i=1,2,\cdots,D$, \\ \quad and $j$ cluster indices $\in \{1,2,\cdots,k\}$.  \\
   5: Given a new FV representation $h_i \in \mathbb{R}^D$, its low dimensional \\ \quad
representation is $\l_i\in \mathbb{R}^{k}$, and its $j$th element \\ \quad
$l_{i,j} = \sum_{p\in clique_{j}}{h_{i,p}}/norm_j$. \\
   \hline

\end{tabular}
}
\label{alg: dr}
\end{table}

\subsubsection{Part III:  Encoding and temporal pyramid pooling layers}

  The encoding and temporal pyramid pooling layers constitute the most important part of our network. It transforms feature representations extracted from a varying number of frames into a fixed-length video-level representation.
Note that these two layers are akin to the encoding and pooling operations in the traditional bag-of-features model. In the traditional bag-of-features model, an image contains a varying number of local features. In order to obtain a fixed-length image representation, one first applies an encoding operation to transform the local feature into a coding vector and then performs pooling to obtain the image level representation. The encoding step has been shown to be essential and pooling with the original local features without encoding usually leads to inferior performance. Similarly, the utilization of the encoding layer FCa in our network is of great importance as well. However, unlike most traditional bag-of-features models, in our work the encoding module FCa is embedded into a deep network structure and its parameters are adapted to the target classification task in an end-to-end fashion. Also, just like using spatial pyramid to incorporate the weak spatial information of local features, here we apply temporal pyramid pooling to better cater for the temporal structure of videos.

In our implementation, we calculate the output of FCa as $\mathbf{Y}_{a}=\sigma (\mathbf{X}\mathbf{W}_{a}+\mathbf{B}_{a})$, where $\mathbf{W}_{a}\in\mathbb{R}^{d\times D}$ and $\mathbf{B}_{a}\in\mathbb{R}^{d\times D}$ are model parameters, $\mathbf{X}\in\mathbb{R}^{n\times d}$ and $\sigma$ denote the input and ``ReLU" activation function respectively. $n$ indicates the number of frames in the input video, $d$ and $D$ are dimensionalities of the input frame-level representation and encoded representation respectively.

\begin{figure}[t]
\begin{center}
\includegraphics[scale=.45]{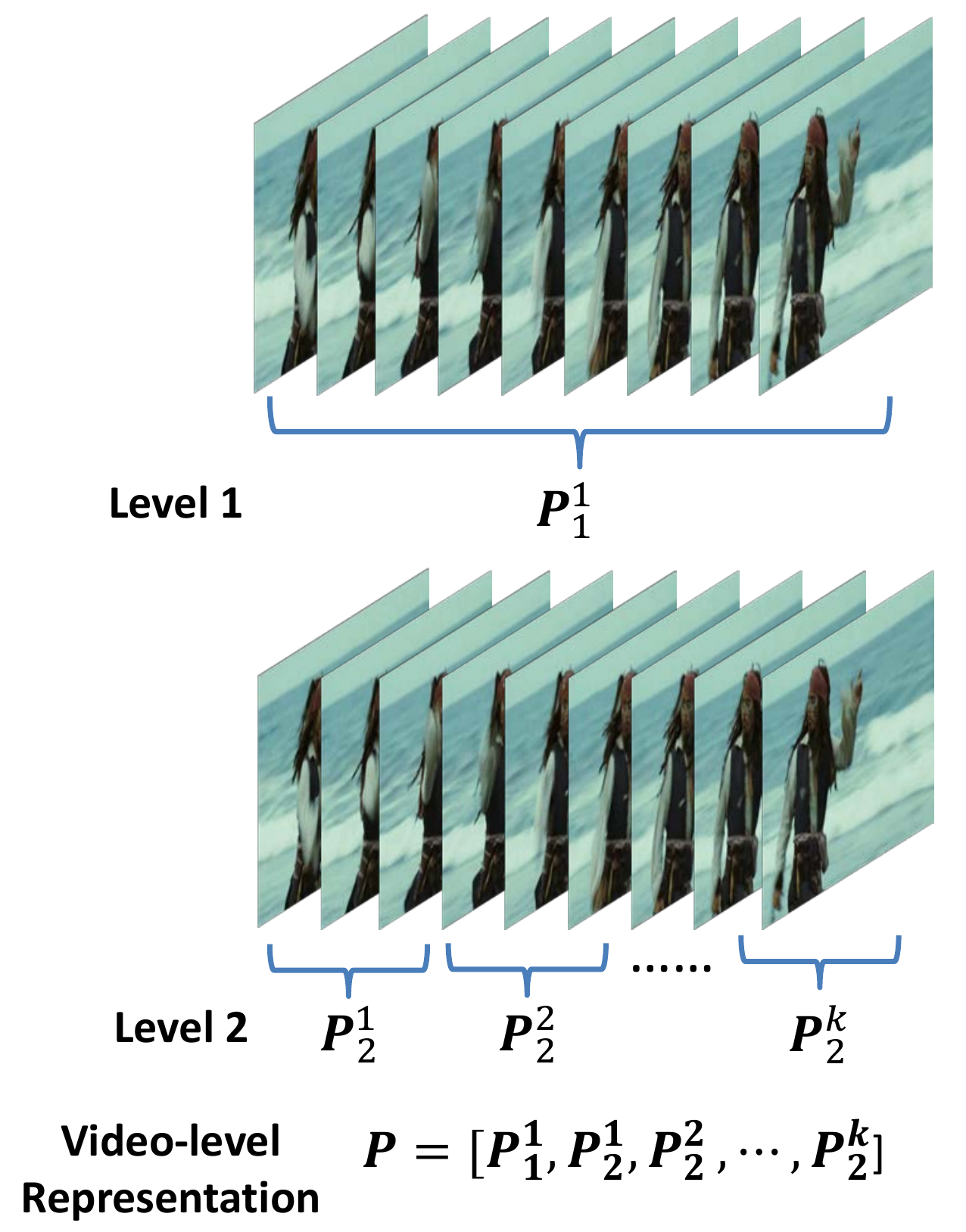}
\end{center}
   \caption{Illustration of temporal pyramid structure in the pooling layer.}
\label{fig:tp}
\end{figure}

  The temporal pyramid pooling strategy is illustrated in Figure \ref{fig:tp}. The input video frames are partitioned in a coarse-to-fine fashion. Here we use two levels of partition. At the coarse level we treat the whole video as a pooling segment. At the fine level we evenly divided the video into multiple segments and perform pooling on each segment. The final video-level representation is obtained by concatenating pooling results from all the segments.

\begin{figure*}[ht]
\begin{center}
\includegraphics[scale=.65]{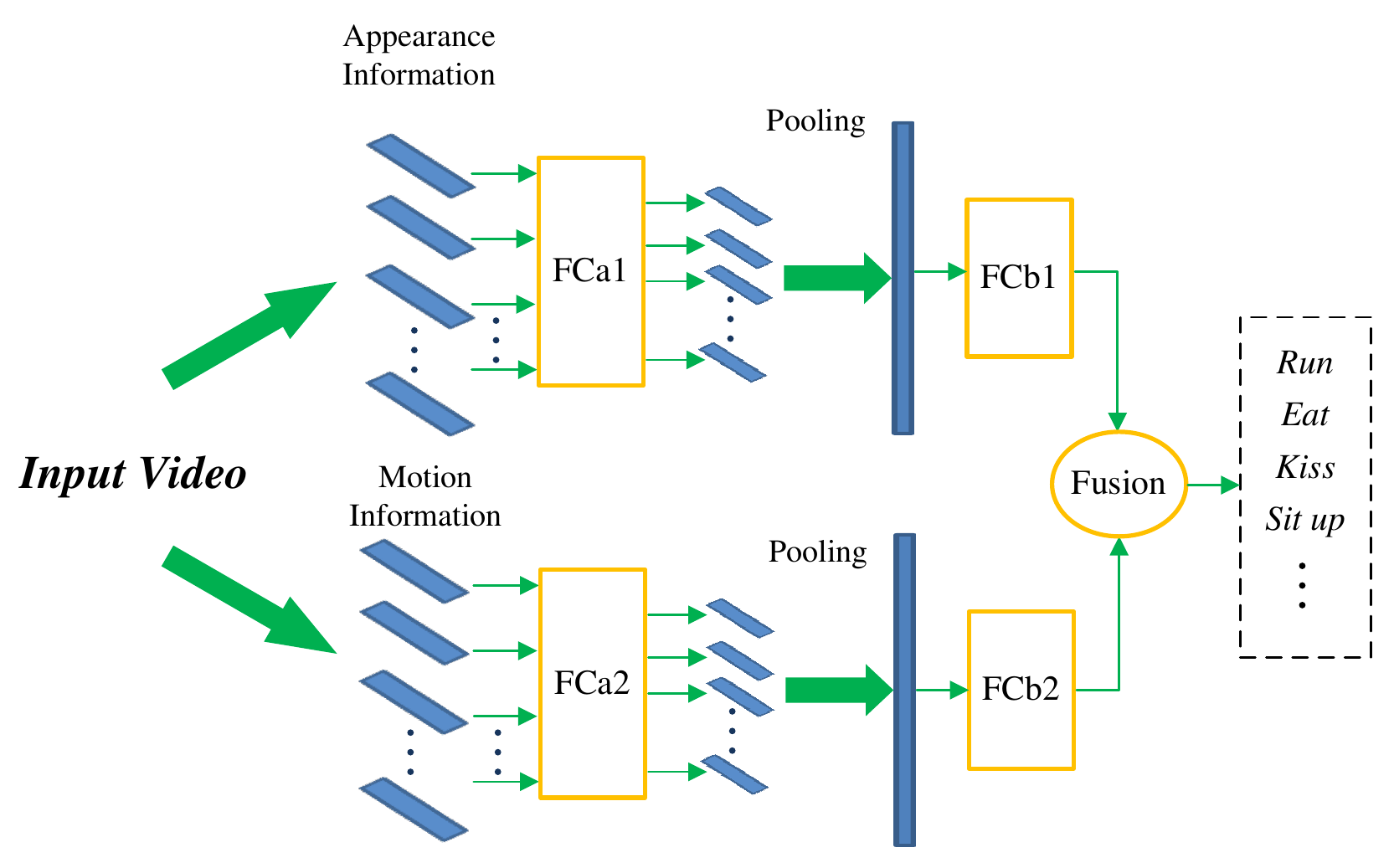}
\end{center}
   \caption{Illustration of late-fusion. We encode, temporally pool the appearance and motion representation separately and fuse the two probability distributions. }
\label{fig:late-fusion}
\end{figure*}

\subsubsection{Part V:  Classification layer}

The final part of our network is a classification layer which classifies the video-level representation obtained from the temporal pyramid pooling layer. It calculates $\mathbf{Y}_{b}=\varphi (\mathrm{pool}\left (\mathbf{Y}_{a}\right )\mathbf{W}_{b}+\mathbf{B}_{b})$ where $\mathbf{W}_{b}$ and $\mathbf{B}_{b}$ are model parameters, $\mathrm{pool}$ and $\varphi$ are pooling and softmax \cite{MatCov} operation respectively. The output $\mathbf{Y}_{b}$ is a probability distribution indicating the likelihood of a video belonging to each class. In the training stage, we use the following loss function to measure the compatibility between this distribution and ground-truth class label:

\begin{align}
	L(\mathbf{W},\mathbf{B})=-\sum_{i=1}^{N}\mathrm{log}(\mathbf{Y}_{b}(c_i)),
\end{align}
where $c_i$ denotes the class label of the $i$th video and $N$ is the total number of training videos. Recall that $\mathbf{Y}_{b}$ is a $\emph{c}$-dimensional vector. Here we use $\mathbf{Y}_{b}(c_i)$ denotes the value at $c_i$th dimension of $\mathbf{Y}_{b}$. Using Stochastic Gradient Descent (SGD), in each step we update model parameters by calculating the gradient of an instance-level loss $L_i(\mathbf{W},\mathbf{B}) = \mathrm{log}(\mathbf{Y}_{b}(c_i))$.

\subsection{Late fusion model}

The aforementioned network structure combines motion and appearance information at the frame level. An alternative way is to fuse these two types of information after obtaining the output of the last layer of our network. We illustrate this scheme in Figure \ref{fig:network}. This scheme consists of two independent network streams. One stream uses appearance information, another stream uses motion information. Each network in these two streams is very similar to that proposed in Figure \ref{fig:network}. The only difference is that the network in Figure \ref{fig:late-fusion} does not have the feature concatenation layer. We independently train these two networks. At the testing stage, we obtain the final output the fused network by calculating the weighted average of $\mathbf{Y}_{b1}$ and $\mathbf{Y}_{b2}$, the outputs from FCb1 and FCb2 respectively.

\subsection{Implementation}
\subsubsection{Motion feature} Our network utilizes both raw frame images and motion features as network input. To calculate the motion feature for a given frame, the Fisher vector encoding is applied to the trajectories falling into its neighbouring 11 frames (from -5 to 5). Following the setting of \cite{Wang2013}, we set the number of Gaussians to 256 for Fisher Vector encoding. While in \cite{Wang2013} each trajectory is composed of five descriptors, including HOG, Trajectory, HOF, MBHx and MBHy, we use only HOF and MBH due to their strong discrimination power. Since the Fisher vector is of high dimensionality, except for Table \ref{tab:overall-hollywood2} and \ref{tab:overall-hmdb51}, the supervised feature merging strategy in Section \ref{motion_featue} is applied to further reduce the frame-level Fisher vector from 76800 dimensions to 4096 dimensions due to the computational reason. Then the input to the network is $n$ 8192-dimensional features where $n$ denotes the number of frames of a video.

\subsubsection{Network training}
In our work, we initialize the parameters of C1-FC7 using a pre-trained model ``\emph{vgg-fast}" \cite{Chatfield14} and keep them fixed during training. During the training procedure, the parameters of FCa and FCb are learned using stochastic gradient descent with momentum. We set momentum to be 0.9 and weight decay to be $5\times{10^{-4}}$. The training includes 25 epochs for all training sets.

\section{Experimental evaluation}
\label{experiment}
\begin{figure*}[ht]
\begin{center}
\captionsetup{justification=centering}
\includegraphics[scale=.7]{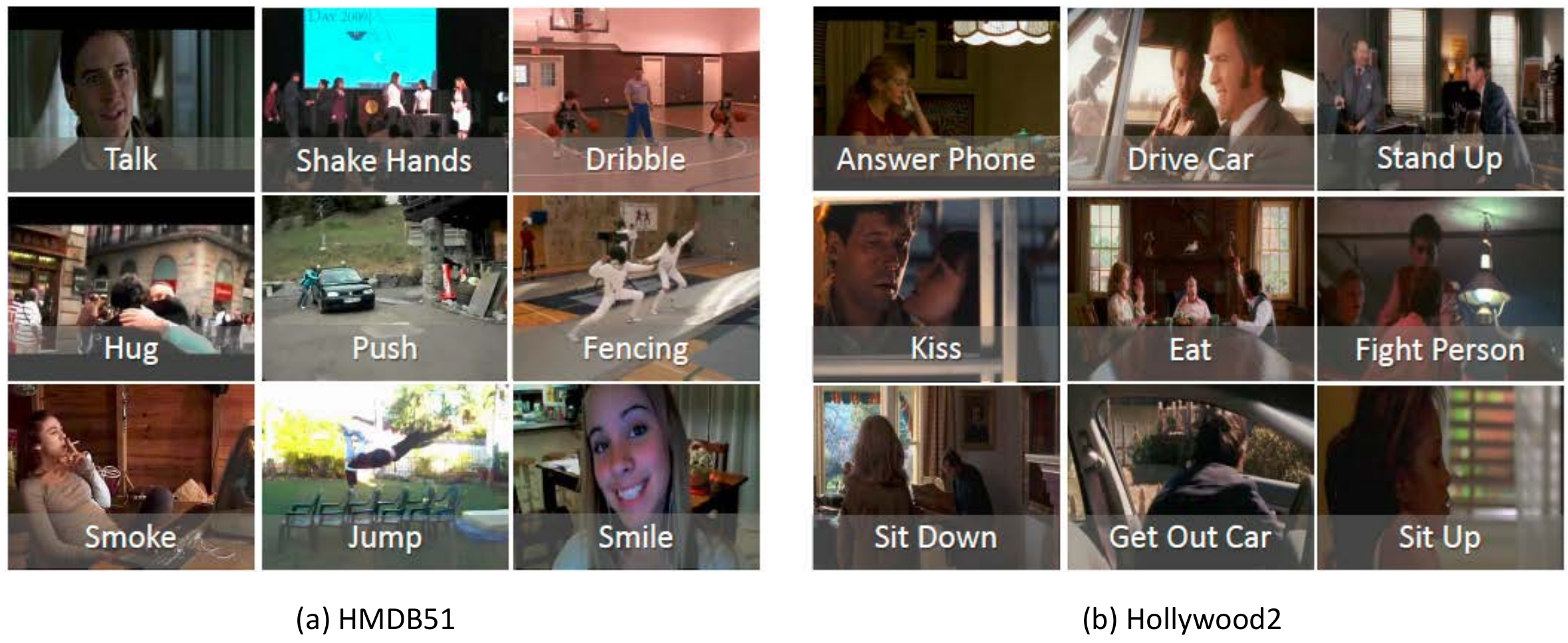}

   \caption{Example frames from (a) HMDB51 and (b) Hollywood2.}
\label{fig:sampled_frms}
\end{center}
\end{figure*}
We conduct a number of experiments on two challenging datasets, Holywood2 and HMDB51 to evaluate the performance of the proposed method and analyze the effects of various components of our network structure. Figure \ref{fig:sampled_frms} give some example frames from HMDB51 and Hollywood2.
\subsection{Experimental setup}
\textbf{Datasets}
 Hollywood2 and HMDB51 can be regarded as two most challenging action recognition datasets because most existing methods achieve very low recognition accuracy on these two datasets. The difficulties lay in that they contain many complex actions and there are a lot of uncontrolled scene variations within videos.

The Hollywood2 dataset \cite{marszalek09} is composed of video clips from 69 movies and includes 12 classes. It contains a total of 1,707 videos with 823 training videos and 884 testing videos. Training and testing videos belong to different movies. The performance is measured by mean average precision (mAP).

The HMDB51 dataset \cite{Kuehne11} is collected from  various sources, such as movies, Prelinger archive and YouTube. It contains 6,766 video clips belonging to 51 classes. According to the protocol in \cite{Kuehne11}, three training-testing splits are provided for this dataset. For each class, there are 70 training videos and 30 testing videos. The average classification accuracy over all the classes and splits is reported. This dataset provides two versions, a stabilized one and an unstabilized one. In our experiments, we use the latter version.

\subsection{Performance evaluation}
In this subsection, we first compare the proposed method to state-of-the-art methods and then discuss three aspects related to our network structure, that is, the evaluation of the `motion part' and the `appearance part' of our network, the comparison of early fusion and late fusion and the effect of temporal pyramid in the pooling layer.

\subsubsection{Hollywood2}

Table \ref{tab:overall-hollywood2} compares our method to several leading methods on this dataset. As can be seen, our method achieves the best performance on this dataset. Compared to improved dense trajectory \cite{Wang2013}, the most competitive one to our method, we have gained more than 4\% improvement. It can also be observed that motion features of high dimensional perform better than low dimensional features.

\begin{table}[htbp]\normalsize
 \caption{Experimental results on Hollywood2. While LD means low dimension (4096) for frame-level motion features, HD indicates high dimension (20000) for frame-level motion features.}
  \centering
  \renewcommand{\arraystretch}{1.1}
    \begin{tabular}{l|c}

    \hline\noalign{\smallskip}
    Dense trajectory \cite{wang:2011} & 58.5\% \\
    Mathe \textit{et al.} \cite{Fitzgibbon} & 61.0\% \\
    Actons \cite{6751554} & 61.4\% \\
    Jain \textit{et al.} \cite{6619174} & 62.5\% \\
    Improved dense trajectory \cite{Wang2013} & 64.3\% \\
    Ours (LD) & \textbf{67.5\%}
    \\
    Ours (HD) & \textbf{68.4\%} \\
    \hline \noalign{\smallskip}
    \end{tabular}%

  \label{tab:overall-hollywood2}%
\end{table}%

\subsubsection{HMDB51}
Table \ref{tab:overall-hmdb51} compares our method to several state-of-the-art methods on HMDB51. As can be seen, our method achieves the second best performance reported on this dataset. Hybrid improved dense trajectories in \cite{peng14}, employs multiple unsupervised encoding methods i.e. Fisher vector \cite{Perronnin:2010}, VLAD \cite{VLAD} and LLC \cite{llc}. In comparison, our method is much more elegant in the sense that it relies on a single encoding module. Note that the best performed method, stacked Fisher vector \cite{peng:stack} employs two-level Fisher vector
encoding and concatenates them together as video representation. If we concatenate global motion encodings and frame-level motion representations (both using four descriptors HOG, HOF, MBHX, MBHY), our performance can be boosted significantly as well.

We also compare our method to the work in \cite{Andrew14} which is also a CNN based method and adopts frame sampling to handle the issue of video-length variation. Our method outperforms it. Note that in their experiment they combine HMDB51 and UCF101 \cite{ucf101} as the training set while our model is trained only on HMDB51. Besides better performance, we believe our network offers a more principled solution to handle the video-length variation issue. Again, we can derive better performance when using high dimensional motion features.

\begin{table}[htbp]\normalsize
\caption{Experimental results on HMDB51. While LD means low dimension (4096) for frame-level motion features, HD indicates high dimension (20000) for frame-level motion features.}
  \centering

  \renewcommand{\arraystretch}{1.1}
  \begin{tabular*}{8cm}{l|c}
    \hline \noalign{\smallskip}
    Spatial-temporal HMAX network \cite{Kuehne11} & 22.8\% \\
    Dense trajectory \cite{wang:2011} & 47.2\% \\
    Jain \textit{et al.} \cite{6619174} & 52.1 \% \\
    Multi-view super vector \cite{6909477} &55.9 \% \\
    Improved dense trajectory \cite{Wang2013} & 57.2\% \\
    Hybrid improved dense trajectories \cite{peng14} & \textbf{61.1\%} \\
    Stacked Fisher Vector \cite{peng:stack} & \textbf{66.8\%} \\
    Two-stream ConvNet (average fusion) \cite{Andrew14} & 58.0\% \\
    Two-stream ConvNet (SVM fusion) \cite{Andrew14} & 59.4\% \\
    Ours (LD)& 59.7\% \\
    Ours (HD) &\textbf{60.8\%}
    \\ \hline \noalign{\smallskip}
    \end{tabular*}%

  \label{tab:overall-hmdb51}%
\end{table}%

\subsection{Fusion of motion net and global Fisher Vector}
In our network structure, we utilize frame-level motion features. To form a frame-level motion representation, we first extract local motion features along trajectories passing this frame and then encode them using Fisher Vector. Different from global Fisher Vector \cite{Wang2013} that embodies global motion via encoding the motion features over the entire video, frame-level motion representation abstracts local semantic information in temporal domain. In this part, we fuse these two kinds of representations together to fully exploit their discriminative power. We adopt score level fusion. For the motion net with motion features as input, we use the outputs of softmax layer. And for global Fisher Vector, we train a linear SVM \cite{REF08a} as in \cite{Wang2013} and outputs probability decision scores. Then these two kinds of scores are fused by averaging. Note that we use HOF and MBH descriptors for both methods.

\begin{table}[htbp]\normalsize
 \caption{Fusion of motion net and global Fisher Vector for motion features.}
        \centering

        \begin{tabular}{lcc}

            \hline\noalign{\smallskip}

                Methods (Dim.)  &    Hollywood2  & HMDB51 \\

            \noalign{\smallskip}

            \hline

            \noalign{\smallskip}

             Global Fisher Vector (76800)   & 62.6\%          &   54.7\%                \\

             Motion Net (4096) & 62.9\%
&	53.4\%				\\
			Score-Level Fusion & \textbf{70.7\%} & \textbf{61.8\%} \\

            \noalign{\smallskip}

            \hline \noalign{\smallskip}

      \end{tabular}

      \label{tab:motion_fusion}

\end{table}

Table \ref{tab:motion_fusion} gives the fusion results. Due to limitation of computational power, we reduce the dimension of frame-level motion representation from 76800 to 4096 to make the network training feasible. However, our network can still achieve comparable performance to high dimensional Fisher Vectors. More importantly, we can see when combining these two kinds of methods together the recognition performance is boosted by a large margin which proves that these two kinds of representations can compensate each other in describing the motions in the videos.
\subsection{Motion vs. appearance}
Our network utilizes both appearance information and motion information. In this section, we test these two sources of information separately to demonstrate their impacts on action recognition. In other words, we discard the feature concatenation layer and only choose one type of feature, motion or appearance as our network input. To further demonstrate the effectiveness of our network structure on each type of feature, we also introduce two baselines which use either motion or appearance information.

\label{experiment}
\begin{figure*}[ht]
\begin{center}
\includegraphics[scale=.7]{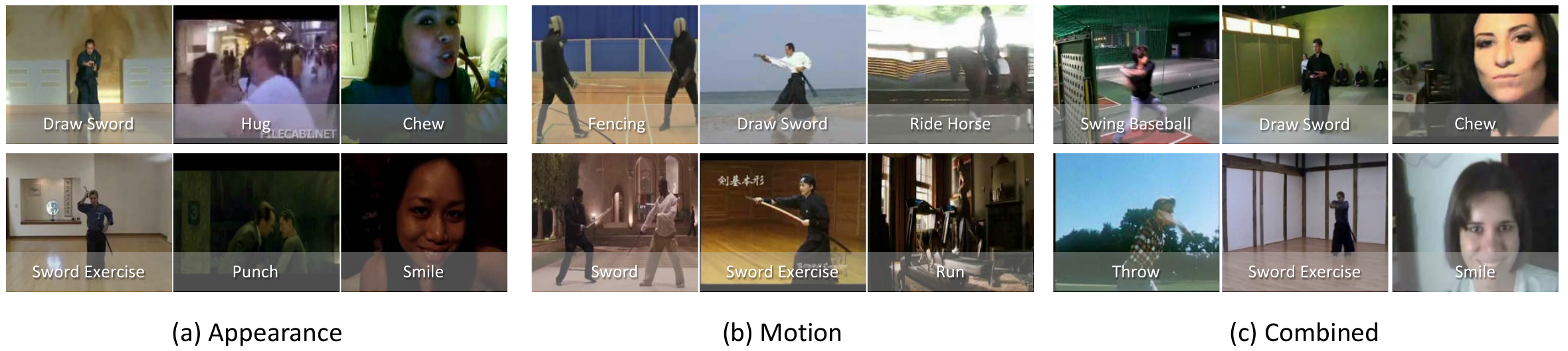}
\end{center}
   \caption{Examples of misclassification for HMDB51 using (a) appearance features, (b) motion features and (c) combined features. The two rows of actions are easily to be misclassified.}
\label{fig:failue}
\end{figure*}

\noindent \textbf{Baseline 1: } The first baseline applies the same CNN that is used for initializing our model parameters C1-FC7 to each frame. The FC7 layer activations are extracted as frame-level features. These features are aggregated through average pooling or temporal pyramid pooling. When temporal pyramid pooling is applied, we use the same pooling parameter as the temporal pyramid pooling layer of our network. After pooling, a linear SVM is applied to classify the pooled video representation. We denote this baseline as Appearance Average Pooling (AAP in short) and its temporal pyramid pooling variant as (ATP).

\noindent \textbf{Baseline 2: } The second baseline adopts frame-level motion feature and creates video-level representation through average pooling or temporal pyramid pooling. To ensure fair comparison, we use HOF and MBH of improved dense trajectory and their dimensionality reduced Fisher vectors as the motion descriptors and frame-level motion feature respectively. We denote this baseline as Trajectory Average Pooling (TAP in short) and its temporal pyramid pooling variant as (TTP).

Table \ref{tab:hollywood2} and \ref{tab:hmdb51} show the results on Hollywood2 and HMDB51. From these two tables we can have the following observations:
\begin{itemize}
	\item The Motion feature contributes more than the appearance feature in achieving good classification performance. From both datasets we can see that motion feature significantly outperforms the appearance feature.
	\item Our network structure works especially well for the appearance feature. In Table \ref{tab:hollywood2}, our method outperforms ATP and AAP by 8\% and 10\% respectively; In Table \ref{tab:hmdb51}, our method outperforms ATP and APP by 1.7\% and 3.8\% respectively. Recall that the major difference between ATP and our network (with the output of FC7 as input) is the encoding layer FCa. The superior performance of our network demonstrates the necessity of applying the encoding layer before pooling.
	\item An interesting observation is that our network structure does not help too much for the motion feature. As can be seen in Table \ref{tab:hollywood2} and Table \ref{tab:hmdb51}, our method achieves comparable performance to TTP, which means that the encoding layer does not lead to too much improvement. This is probably because the frame-level motion feature we used is already a coding vector, Fisher vector namely, and it is ready for the pooling operation. Thus, adding another encoding layer will not bring too much improvement. In contrast, the output of FC7 is not well-tuned for pooling (recall that the layers before FC7 is pretrained on a CNN without pooling layer), thus adding the encoding layer is beneficial.
	\item Finally, we observe that adding temporal pyramid into feature pooling can obviously improve the classification performance since it can better describe the temporal structure of videos.

\end{itemize}

Figure \ref{fig:failue} give some failed examples for HMDB51. As can be seen, these actions tended to be misclassified are very similar w.r.t both appearances and motion patterns.

\begin{table}[h]\normalsize
\caption{Comparison of the network to baselines on Hollywood2 using appearance information or motion information.}
  \centering
  \renewcommand{\arraystretch}{1.2}
    \begin{tabular}{l|p{1.5cm}c}
    \hline
    \multirow{3}{*}{Appearance} & AAP & 34.2\% \\
    & ATP & 36.3\% \\
    & Ours & \textbf{44.4\%}
    \\ \hline \hline
	\multirow{3}{*}{ Motion} & TAP & 60.3\% \\
    & TTP & 62.7\% \\
    & Ours & \textbf{62.9\%} \\
    \hline \noalign{\smallskip}
    \end{tabular}%

  \label{tab:hollywood2}%
\end{table}%

\begin{table}[h]\normalsize
 \caption{Comparison of the network to baselines on HMDB51 using appearance information or motion information.}
  \centering
  \renewcommand{\arraystretch}{1.2}
    \begin{tabular}{l|p{1.5cm}c}
    \hline
    \multirow{3}{*}{Appearance} & AAP & 37.5\% \\
    & ATP & 39.6\% \\
    & Ours & \textbf{41.3\%}
    \\ \hline \hline
	\multirow{3}{*}{ Motion} & TAP & 50.9\% \\
    & TTP & 53.5\% \\
    & Ours & 53.4\%
    \\ \hline \noalign{\smallskip}

    \end{tabular}%

  \label{tab:hmdb51}%
\end{table}%

\subsection{Early fusion vs. late fusion}

In this part, we compare two types of fusion methods, namely early fusion and late fusion. While early fusion concatenates motion features and appearance features together as input to train the network, late fusion, as in \cite{Andrew14}, trains two independent networks using appearance feature and motion feature separately and combines the softmax scores by simple weighted averaging. As can be seen from Table \ref{tab:fusion}, both fusion methods boost the classification performance comparing to network with single input, which proves that appearance information and motion information are complementary. Further, early fusion obviously outperforms late fusion, improving the results by around 3\% and 2\% on Hollywood2 and HMDB51 respectively. Also, we show the average precision for each class of Hollywood2 in Table \ref{tab:ap_hollywood2}. We can see that apart from ``Eat", early fusion performs better on all the other actions. The reason may lie in that we train the motion stream and appearance stream independently in the late fusion model without adapting model parameters towards optimal combination of two types of information.

\begin{table}[htbp]\normalsize

      \caption{Comparison between early fusion and late fusion. For late fusion, we use $1/3$ appearance weight and $2/3$ motion weight.}

        \centering

        \begin{tabular}{lcc}

            \hline\noalign{\smallskip}

                Methods  &    Hollywood2  & HMDB51 \\

            \noalign{\smallskip}

            \hline

            \noalign{\smallskip}

             Late Fusion   & 64.7\%          &   57.7\%                \\

             Early Fusion

& \textbf{67.5\%}       &   \textbf{59.7\%}                \\

            \noalign{\smallskip}

            \hline \noalign{\smallskip}

      \end{tabular}

      \label{tab:fusion}

\end{table}

\begin{table*}[htbp]\normalsize

      \caption{Average Precisions (AP) for each class of Hollywood2. LF represents late-fusion and EF represents early-fusion.}

        \centering

        \begin{tabular}{lcccccccccccc}

            \hline\noalign{\smallskip}

              & AnswerPhone & DriveCar & Eat & FightPerson & GetOutCar & HandShake \\

            \noalign{\smallskip}

            \hline

            \noalign{\smallskip}

             LF & 49.1\% & 94.1\% & 58.7\% & 69.0\% & 81.8\% &40.3\%  \\

             EF &49.8\% &95.3\% & 54.1\% & 69.4\% & 82.0\% & 51.4\% \\

            \noalign{\smallskip}

            \hline\hline

            \noalign{\smallskip}

&  HugPerson & Kiss & Run & SitDown & SitUp & StandUp \\

\noalign{\smallskip} \hline \noalign{\smallskip}

LF & 45.8\% & 62.6\% & 86.1 \% & 76.2\% & 32.6\% &79.8\%      \\

EF &  55.1\% & 66.3\% & 88.2\% & 76.7\% & 36.9\% & 84.7\%           \\

\noalign{\smallskip}

            \hline

            \noalign{\smallskip}

      \end{tabular}

      \label{tab:ap_hollywood2}

\end{table*}

\subsection{The impact of the temporal pyramid parameter}
In this subsection, we evaluate the effects of temporal pyramid in the pooling layer. Intuitively, adding temporal-pyramid can better cater for the video structure. As in previous experiments, we choose a two-level temporal pyramid structure with one level covering all video frames and another level dividing a video into $b$ segments. Here we evaluate the impact of $b$. We vary $b$ from 0 to 7, where $b=0$ means that no temporal pyramid is utilized. To simplify the experiment, we only conduct experiments on appearance features. As can be seen in Table \ref{tab:tp-pooling}, adding more segments significantly boosts the results initially and we achieve best performance at $b=5$. After that peak point continuing to add segments will lead to worse results.

\begin{table}[t]\normalsize
 \caption{The impact of the value of $b$ in the temporal pyramid pooling layer.}
  \centering
  \renewcommand{\arraystretch}{1.2}
    \begin{tabular}{l|cc}
    \hline
    \multirow{4}{*}{Hollywood2} & $b=0$ & 38.2\% \\
    & $b=3$ & 44.1\% \\
    & $b=5$ & 44.2\% \\
    & $b=7$ & 43.9\% \\
    \hline \hline
	\multirow{4}{*}{ HMDB51} & $b=0$ & 38.5\% \\
    & $b=3$ & 39.4\% \\
    & $b=5$ & 41.3\% \\
    & $b=7$ & 39.6 \%\\
    \hline \noalign{\smallskip}

    \end{tabular}%

  \label{tab:tp-pooling}%
\end{table}%

\section{Conclusions}
\label{conclusion}
We propose a deep CNN structure which allows a varying number of video frames as network input and apply this network to action recognition. This model achieves superior performance while requiring fewer training videos. It also enables us to combine the appearance feature from a CNN and the state-of-the-art motion descriptor.

\ifCLASSOPTIONcaptionsoff
  \newpage
\fi

\bibliographystyle{IEEEtran}
\bibliography{CSRef}

\end{document}